\documentclass[journal]{IEEEtran}
\ifCLASSINFOpdf
\else
\fi
\usepackage{graphicx}
\usepackage{epstopdf}
\usepackage{subfigure}
\usepackage{amssymb,amsmath}
\usepackage{amsfonts}
\usepackage[ruled]{algorithm2e}
\usepackage{algpseudocode}
\usepackage{multirow}
\usepackage{multicol}
\usepackage[flushleft]{threeparttable}
\usepackage{balance}
\usepackage{url}

\begin{document}
%
\title{Semantic Softmax Loss for Zero-Shot Learning}
%
%
%

\author{Zhong~Ji,~\IEEEmembership{Member,~IEEE,}
        Yunxin~Sun,
        Yulong~Yu*,
        Jichang~Guo,
        and Yanwei Pang, \IEEEmembership{Senior Member,~IEEE}
\thanks{This work was supported by the National Basic Research Program of China (973 Program) under Grant 2014CB340400, the National Natural Science Foundation of China under Grants 61472273 and 61632018.}
\thanks{Z.~Ji, Y.~Sun, Y.~Yu* (corresponding author), J.~Guo and Y.~Pang are with the School of Electrical and Information Engineering, Tianjin University, Tinjin 300072, China (e-mails: \{jizhong, sunyuxin, yuyunlong, jcguo, pyw\}@tju.edu.cn).}
}

\maketitle

\begin{abstract}
     A typical pipeline for Zero-Shot Learning (ZSL) is to integrate the visual features and the class semantic descriptors into a multimodal framework with a linear or bilinear model. However, the visual features and the class semantic descriptors locate in different structural spaces, a linear or bilinear model can not capture the semantic interactions between different modalities well. In this letter, we propose a nonlinear approach to impose ZSL as a multi-class classification problem via a Semantic Softmax Loss by embedding the class semantic descriptors into the softmax layer of multi-class classification network. To narrow the structural differences between the visual features and semantic descriptors, we further use an $L_2$ normalization constraint to the differences between the visual features and visual prototypes reconstructed with the semantic descriptors. The results on three benchmark datasets, i.e., AwA, CUB and SUN demonstrate the proposed approach can boost the performances steadily and achieve the state-of-the-art performance for both zero-shot classification and zero-shot retrieval.
\end{abstract}

%
\IEEEpeerreviewmaketitle

\section{Introduction}
%
%
%
%
Zero-Shot Learning (ZSL) \cite{Lampert09cvpr,Shaoling17spl,Akata15cvpr,Fu15pami,Yu17arXiv,Simonyan15iclr} aims at building classifiers to predict the unseen classes without any visual instances in the training stage. This task is achieved by transferring the information from seen classes to unseen ones with the knowledge about how each unseen class is semantically related to the seen classes. In order to measure the semantic relations between different classes, both the seen classes and unseen ones are represented as a high dimensional vector embedded in a semantic space. Such a space can be semantic attribute space or semantic word vector space.

Most of the existing ZSL approaches address this task as two different independent subtasks, which can be divided into two categories. The first one associates attribute prediction followed by classification inference \cite{Lampert09cvpr,Farhadi09cvpr,Yu10eccv}. One of the most popular among these approaches is the direct attribute prediction (DAP) approach \cite{Lampert09cvpr}, which predicts attributes independently using SVMs and infers zero-shot predictions by a maximum a posteriori rule that assumes attribute independence. The other one decomposes ZSL into a multimodal learning process and a similarity measurement process. To construct the interactions between the visual instances and the class semantic descriptors, exiting approaches either project the features from one modality to another \cite{Ji17ins,Lazaridou14acl,Socher13nips} or project the features from both modalities into a common space \cite{Akata15cvpr,Zhang16cvpr,Romera-Paredes15icml,Fu15pami,Yu17arXiv}. To measure the similarity, most approaches use nearest neighbour classifier (NN) \cite{Lampert09cvpr,Akata15cvpr,Zhang16cvpr} or label propagation \cite{Kodirov15iccv}.

Although existing approaches for ZSL have achieved impressive performances, they still suffer from issues below. (1) Most existing methods use a linear or bilinear approach to train the multimodal learning model that may not capture the semantic interactions between different modalities well. (2) Existing approaches perform ZSL as two disjoint subtasks, which leads to the information loss.

In this work, we present an end-to-end nonlinear embedding paradigm for ZSL based on the multi-class classification, as illustrated in Fig.1. Specifically, we embed the class semantic descriptors into a multi-class classification framework with the proposed Semantic Softmax Loss (SSL). It divides the classifier parameters into two matrices, a learned generative matrix and an off-the-shelf class semantic matrix. In this way, the visual instances, class semantic descriptors and the class labels are formulated into a unified multi-class classification model, which can be trained in an end-to-end way. We call the proposed method for ZSL as SSL-ZSL for short. Besides, the classification parameters for each class can be seen as a visual prototype reconstructed by the corresponding class semantic descriptor. We impose an $L_2$ normalization constraint to reconstruction task for semantic embedding so that the reconstructed prototypes preserve most of the information.


In summary, this paper contributes to the following aspects:
\begin{itemize}
  \item  We propose an end-to-end framework for ZSL by embedding the class semantic descriptors into the softmax layer in a multi-class classification pipeline, in which the compatibility between the class semantic descriptors and visual instances are optimized under the supervision of labels. In this way, the classifiers of unseen classes can be obtained with the semantic descriptors.
  \item  To narrow the structural differences between the visual and the class semantic spaces, we add an $L_2$ normalization constraint on the visual features and reconstructed visual prototypes such that they lie on the same hypersphere.
  \item  The performances of the proposed approach yield a consistent and significant boost on three benchmark ZSL datasets, namely AwA, CUB and SUN.
\end{itemize}

\begin{figure}[t]
\begin{center}
   \includegraphics[width=1.0\linewidth]{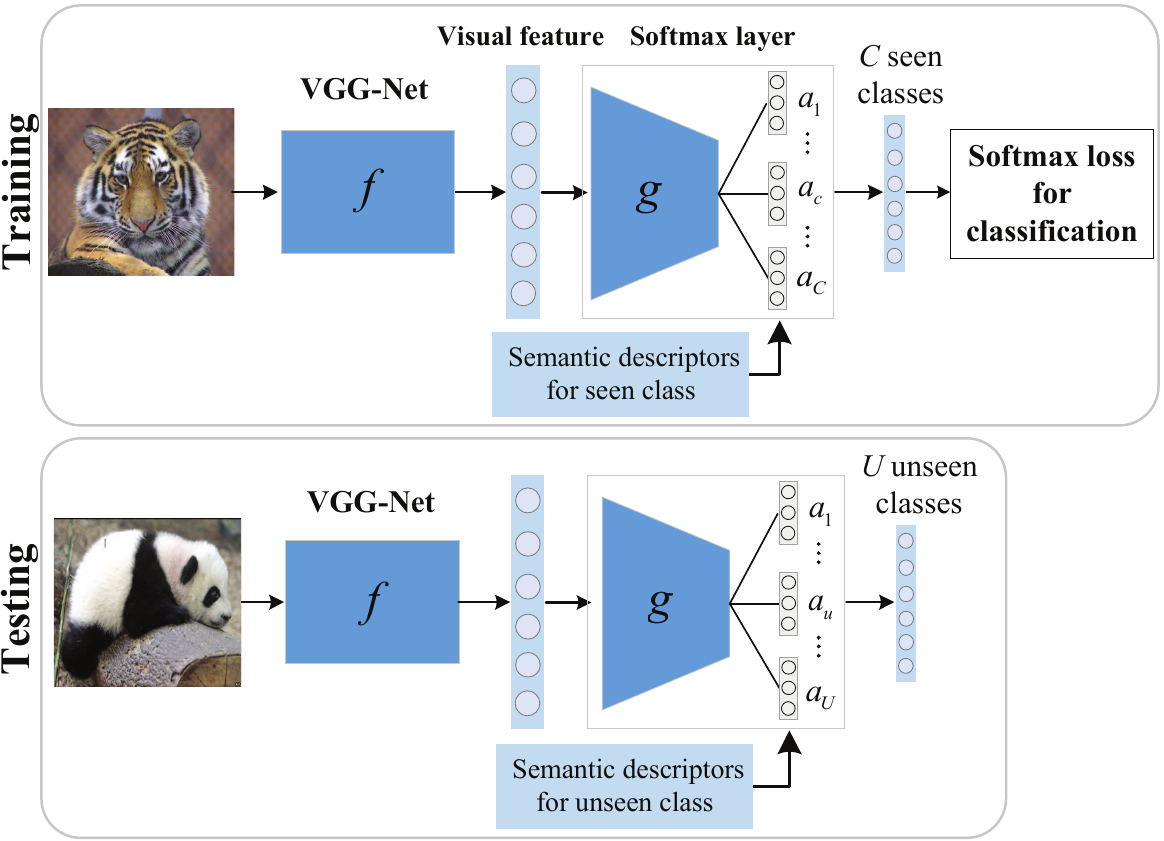}
\end{center}
   \caption{The proposed pipeline for zero-shot learning. In the training stage, the images and the class semantic descriptors from seen classes are taken as input to predict the class label. The VGG-Net is adopted to extract the visual feature, and the class semantic descriptors are embedded in the softmax layer, $f$ and $g$ are models to be trained.
   In the testing stage, the test image and the class semantic descriptors of all candidate unseen classes are taken as input, and outputs the predicted label.}
\label{fig:long}
\label{fig:onecol}
\end{figure}

\section{Semantic Softmax Loss for Zero-Shot Learning}
Given a training dataset with $M$ images and their corresponding labels, a traditional classification model is trained to classify a given image to its correct label. In a typical convolutional neural network (CNN), a softmax loss function is commonly used for training the network, given by Eq.(1)
\begin{equation}
  L_S = -\frac{1}{M}\sum_{i=1}^{M}\log{\frac{e^{W_{y_i}^{T}f(\mathbf{x}_i)+b_{y_i}}}{\sum_{j=1}^{C}e^{W_j^{T}f(\mathbf{x}_i)+b_j}}},
\end{equation}
where $C$ is the number of training classes, $\mathbf{x}_i$ is the $i^{th}$ instance. In the softmax loss,  $f(\mathbf{x}_i)$ is usually the corresponding output of the penultimate layer of a CNN, $y_i$ is the corresponding class label, and $W$ and $b$ are the weights and bias for the last layer of the network which act as a classifier.

For the seen classes, the classifier parameters $\{W,b\}$ can be obtained by training the network with training instances. For the unseen classes, however, no instances are available for training the classifiers. Consequently, this pipeline can not be applied for ZSL directly.

To address the above issue, existing ZSL approaches introduce the class semantic descriptors to transfer the knowledge from seen classes to unseen ones. Considering that the class semantic descriptor characterizes the properties of a class, it is reasonable to assume that the class classifier can be derived from its corresponding class semantic descriptor,
\begin{equation}
  W_j = g(\mathbf{a}_j),
\end{equation}
where $\mathbf{a}_j$ and $W_j$ are the semantic descriptor and classifier for $j^{th}$ class, respectively. Function $g(\cdot)$ denotes the mapping between the class semantic descriptors and classifiers, which can be linear or nonlinear function. In this paper, we consider a simple linear model, i.e., $W_j=V^T\mathbf{a}_j$.

With the learned mapping function $V$, a class classifier can be deduced by Eq.(2) from its class semantic descriptor. Subsequently, ZSL can be performed as follows:

(1). The visual feature $f(\mathbf{x}_i)$ of a test image $\mathbf{x}_i$ from an unseen class is first extracted using the pre-trained CNN, and normalized to unit length.

(2). The classification of $\mathbf{x}_i$ is achieved by calculating the compatibility scores of the visual feature $f(\mathbf{x}_i)$ and the classifiers of all candidate unseen classes:
\begin{equation}
  c(\mathbf{x}_i) = \arg\max_j{S(f(\mathbf{x}_i),W_j)},
\end{equation}
where $S(f(\mathbf{x}_i),W_j)$ denotes the compatibility score between the instance $\mathbf{x}_i$ and class $j$.

Usually, the compatibility score is computed with the inner product between the two vectors or by adapting the cosine similarity as given by Eq.(4):
\begin{equation}
  S(\mathbf{x}_i,W_j) = \frac{f(\mathbf{x}_i)^{T}W_j}{\|f(\mathbf{x}_i)\|_2\|W_j\|_2}.
\end{equation}

From another view, $W_j$ in Eq.(2) can be seen as the visual prototype that is reconstructed by the class semantic descriptor. Since each class is represented as a single semantic descriptor which is insufficient to fully represent what that class looks like. Consequently, even if the reconstructed visual prototype is learned by enforcing the class semantic descriptor to be projected to the center among its visual instances in the visual space, the classifiers will still struggle to assign the correct class labels.

To solve this issue, we impose the $L_2$-norm of the features to be fixed for every training instance as well as the reconstructed visual prototypes. Specifically, we add an $L_2$ normalization constraint to the substraction of the visual features and the reconstructed visual prototypes such that they lie on the same hypersphere. This approach has two advantages. Firstly, on a hypersphere, minimizing the softmax loss is equivalent to maximizing the cosine similarity for the positive pairs and minimizing it for the negative pairs, which strengthens the discriminative ability of the classifier. Secondly, the softmax loss is able to narrow the information loss caused by the different structure from different modalities, since all the features have same $L_2$-norm.

The proposed semantic softmax loss is given by Eq.(5).
\begin{equation}\
    \begin{aligned}
        &L_S = -\frac{1}{M}\sum_{i=1}^{M}\log{\frac{e^{\mathbf{a}_{y_i}^{T}Vf(\mathbf{x}_i)+b_{y_i}}}{\sum_{j=1}^{C}e^{\mathbf{a}_j^{T}Vf(\mathbf{x}_i)+b_j}}} + \lambda\|V\|_F^{2},\\
        &\emph{s.t.}  \ \ \ \|f(\mathbf{x}_i) - V^T\mathbf{a}_{y_i}\|_2=\alpha, \ \ \forall_i=1,2,...,M, \\
    \end{aligned}
\end{equation}
where $\{V,b\}$ are the parameters to be trained, $\|V\|_F^{2}$ is the regularizer, and $\lambda$ is the constant hyper-parameter.



Here, we provide the details of implementing the semantic embedding in Eq.(5) in the framework of multi-class classification. This module is added just after the penultimate layer of CNN which acts as a feature descriptor. The class semantic descriptors are embeded into the softmax layer based on the inner product. The parameters for softmax layer are derived from the class semantic descriptors and a compatible matrix that shared across all classes. Meanwhile, the difference between the visual feature and the reconstructed visual prototype are scaled to a hypersphere of a fixed radius with an $L_2$-normalization layer.

Fixed $\alpha$ as a constant, the $L_2$ normalization constraint is added as a regularizer to the loss function. In this way, the module is fully differentiable and can be used in an end-to-end training of the network. During training, we need to back-propagate the gradient of loss $L_S$ through this module as well as the gradient with respect to the parameters $\{V,b\}$ in the proposed module. If the visual feature is abstracted with the pre-trained CNN in advance, $\{V,b\}$ are the only parameters to be trained. 
\section{Experiments}
In this section, we conduct zero-shot classification and zero-shot retrieval on three benchmark datasets, respectively, and compare the proposed approach with a number of ZSL approaches. We will show the superior performances of our approach against a number of state-of-the-art methods.
\subsection{Datasets and Settings}
\textbf{Datasets.} Three datasets are chosen for our evaluations, Animal with Attributes (AwA) \cite{Lampert09cvpr}, Caltech UCSD Birds (CUB) \cite{Catherine11} and SUN attribute dataset \cite{Genevieve14ijcv}. AwA provides 30,475 images from 50 animal classes, and 85 associated class-level attributes. We follow the standard seen/unseen split \cite{Lampert09cvpr}, where 40 classes with 24,295 images are taken as the seen domain and the remaining 10 classes with 6180 images are adopted as the unseen domain. CUB dataset contains 11,788 images from 200 bird species with 312 associated attributes. In this dataset, we use the same zero-shot split as \cite{Akata15cvpr} with 150 classes for seen data and 50 disjoint classes for the unseen data. SUN dataset contains 717 scene categories annotated by 102 attributes, and each class has 20 images. In this dataset, we use 707 classes as the seen domain and the remaining 10 classes as the unseen domain, the same as that in \cite{Zhang16cvpr}.\\
\textbf{Visual features and class semantic descriptors.} To extract the visual feature for each image, we use the pre-trained VGG-Verydeep-16 model \cite{Simonyan15iclr}, where the output of the penultimate layer (before the softmax) is taken as the feature vector. With regard to class semantic descriptors, we use not only the class attributes associated with the datasets but also the word embeddings for each class.  We train a word2vector \cite{Mikolov13nips} model on the Wikipedia corpus to obtain the 1000-dimensional word vector for each class name. Since few competitors use word embeddings for SUN dataset, we only extract the word vectors for AwA and CUB datasets for the experimental comparison.\\

\subsubsection{Zero-shot Classification}
For zero-shot classification, the model is first trained with the seen data, and then the test images are predicted to the candidate unseen classes with the trained model.\\
\textbf{Competitors.}
We compare our proposed approach with 8 state-of-the-art approaches below:\\
\begin{enumerate}
  \item  \textbf{LR} \cite{Lazaridou14acl} and \textbf{RLR} \cite{Shigeto15mlkd}. As a baseline method, Linear Regression (LR) \cite{Lazaridou14acl} learns a mapping function to project the visual feature to the class semantic space. To alleviate the hubness problem that suffered by nearest neighbor search in a high dimensional space, Reverse Linear Regression (RLR) \cite{Shigeto15mlkd} learns a reverse projection to project the class semantic descriptors to the visual space.
  \item \textbf{ESZSL} \cite{Romera-Paredes15icml} and \textbf{SJE} \cite{Akata15cvpr}. Embarrassingly Simple ZSL (ESZSL) \cite{Romera-Paredes15icml} is a simple but effective approach that integrates the compatibility scores and class labels into a linear framework, where the compatibility scores are the similarities between the visual feature and class semantic descriptor obtained with a bilinear formulation. Likewise, Structured Joint Embedding (SJE) \cite{Akata15cvpr} also uses bilinear compatibility function to associate the visual and class semantic descriptors and adopts a weighted approximate ranking loss inspired from the structured SVM \cite{Tsochantaridis05jmlr}.
  \item \textbf{SSE} \cite{Zhang15iccv} and \textbf{JLSE} \cite{Zhang16cvpr}. Semantic Similarity Embedding (SSE) \cite{Zhang15iccv} and Joint Latent Similarity Embedding (JLSE) \cite{Zhang16cvpr}  express visual images and class semantic descriptors as a mixture of seen class proportions. Specifically, SSE leverages the similar class relationships both in visual and class semantic space and JLSE poses both the visual images and class semantic descriptors into a latent space where the semantic information matches.
  \item \textbf{MLZSC} \cite{Bucher16eccv} and \textbf{MCME} \cite{Ji17ins}. Metric Learning for Zero-Shot Classification (MLZSC) \cite{Bucher16eccv} formulates zero-shot classification as a metric learning problem via improving semantic embedding consistency. Manifold Regularized Cross-Modal Embedding (MCME) \cite{Ji17ins} improves the cross modal embedding ability with an effective manifold regularizer.
\end{enumerate}
\begin{table}
\caption{\upshape{Results on three benchmark datasets in average per-class top-1 accuracy (\%). We compare with approaches under different class semantic descriptors including attributes (A) and word vectors (W). `$\dagger$' denotes the methods are implied by ourselves. `-' indicates that no experiments have been performed under this case in original paper. }}
    \label{Table.2}
    \begin{center}
    \begin{tabular}{|c|c|c|c|c|c|}
    \hline
       \multirow{2}{*}{Method} &\multicolumn{2}{c|}{AwA} &\multicolumn{2}{c|}{CUB}&SUN\\
    \cline{2-6}
     &A &W &A &W & A\\
    \hline
    LR$\dagger$ \cite{Lazaridou14acl} &63.6 &50.6 &37.4 &28.8 &75\\
    RLR$\dagger$ \cite{Shigeto15mlkd} &73.7 &58.4 &35.2 &26.5&76\\
    SSE \cite{Zhang15iccv} &76.3 & - & 30.4 & - & 82.5\\
    SJE \cite{Akata15cvpr} & 66.7 & 51.2 & 50.1 & 28.4 &- \\
    ESZSL$\dagger$ \cite{Romera-Paredes15icml} &76.5 &71.5 &47.6 &30.9&82.0\\
    JLSE \cite{Zhang16cvpr} & 80.5 &- &41.8 & - & 83.8 \\
    MLZSC \cite{Bucher16eccv} &77.3 &- &43.3 &- &84.4\\
    MCME \cite{Ji17ins} & - &67.0 & - & 32.6 &- \\
    \hline
    SSL-ZSL &\textbf{82.69} &\textbf{72.02}& \textbf{55.72} &\textbf{33.33} &\textbf{88.00}\\
    \hline
    \end{tabular}
\end{center}
\end{table}
\textbf{Evaluation Criteria.}
We average the correct prediction independently for each class before dividing the number of classes, i.e., the average per-class top-1 accuracy, which is popular for zero-shot classification.\\
\textbf{Comparison Results.}
Table \uppercase\expandafter{\romannumeral1} presents the comparative results of SSL-ZSL on three datasets. It is worth mentioning that  SJE \cite{Akata15cvpr} extracts GoogleNet features as image representations, the others all use VGG features. From the results, we can observe that our proposed approach achieves the best results on all datasets. Specifically, it has an impressive gains over the other state-of-the-art methods ranging from 0.52\% to 5.62\% in different datasets with different class semantic descriptors. Besides, the proposed approach has overwhelming superiority than ESZSL \cite{Romera-Paredes15icml}, which is a similar approach with linear model. This indicates the superiority of our proposed nonlinear model.\\
\textbf{The impact of $L_2$ normalization constraint.}
We also conduct experiments to verify the effectiveness of the $L_2$ normalization constraint. We list the classification results of with and without $L_2$ normalization constraint in Table \uppercase\expandafter{\romannumeral2}. We can find that the $L_2$ normalization constraint has large gains of on three datasets. More specifically, it brings 1.77\%, 4.81\% and 1.5\% improvements on AwA, CUB and SUN with attributes, respectively. For displayed directly, we also give a visualization of unseen instances from AwA dataset with and without $L_2$ normalization constraint, as illustrated in Fig.~2. As we can see, the reconstructed visual prototypes with $L_2$ normalization constraint are closer to the centers of respective classes than those of without $L_2$ normalization constraint.
\begin{table}
\caption{\upshape{The classification performances (\%) with and without $L_2$ normalization constraint on three datasets. SSL-ZSL/with and SSL-ZSL/without denote the methods with and without $L_2$ normalization constraint, respectively.}}
\begin{center}
\begin{tabular}{|l|c|c|c|}
\hline
Method & AwA &CUB & SUN  \\
\hline\hline
SSL-ZSL/without  & 80.92 & 50.91 & 86.5 \\
SSL-ZSL/with & 82.69 & 55.72 & 88.0 \\
\hline
Improvement & \textbf{1.77} & \textbf{4.81} & \textbf{1.5} \\
\hline
\end{tabular}
\end{center}
\end{table}
\subsubsection{Zero-shot Retrieval}
Given a specified class semantic descriptors of unseen classes, the task of Zero-shot Retrieval (ZSR) is to search some visual images from an image database related to it. In the experiment, the model is first trained with the seen instances and the class semantic descriptors of unseen classes are then taken as queries to rank the images from unseen classes based on the similarity with the specified query.

We select three existing state-of-the-art ZSR approaches which are published in the past two years for comparison. Table \uppercase\expandafter{\romannumeral3} presents the comparative results for mAP on three benchmark datasets. We can find that the proposed approach achieves the best performances on all datasets. Specifically, SSL outperforms the state-of-the-art methods in 5.52\%, 15.79\% and 6.14\% on AwA, CUB and SUN datasets, respectively. Besides, the proposed SSL achieves 68.23\% average on three datasets, which has a 9.29\% gain than the runner up \cite{Zhang16cvpr}. We argue that the superior performances benefit from our proposed effective optimization model that narrows the structural differences between the visual space and class semantic space.

Fig.3 shows the precision-recall curves for unseen classes on three datasets. Specifically, we provide the precision-recall curves with attribute and word vectors for AwA dataset.
As to CUB dataset, we only show the first 10 classes from 50 unseen classes for the convenience of display. Compared with the precision-recall curves in original paper \cite{Bucher16eccv}, our approach obviously performs superior for most classes on AwA and SUN datasets and has a larger area under the curves for CUB dataset. We also can find that the areas under the curves on AwA dataset are larger than those on CUB and SUN datasets. This is because CUB and SUN are fine-grained datasets which are more challenging than AwA dataset.

\begin{figure}[t]
\begin{center}
   \includegraphics[width=0.99\linewidth]{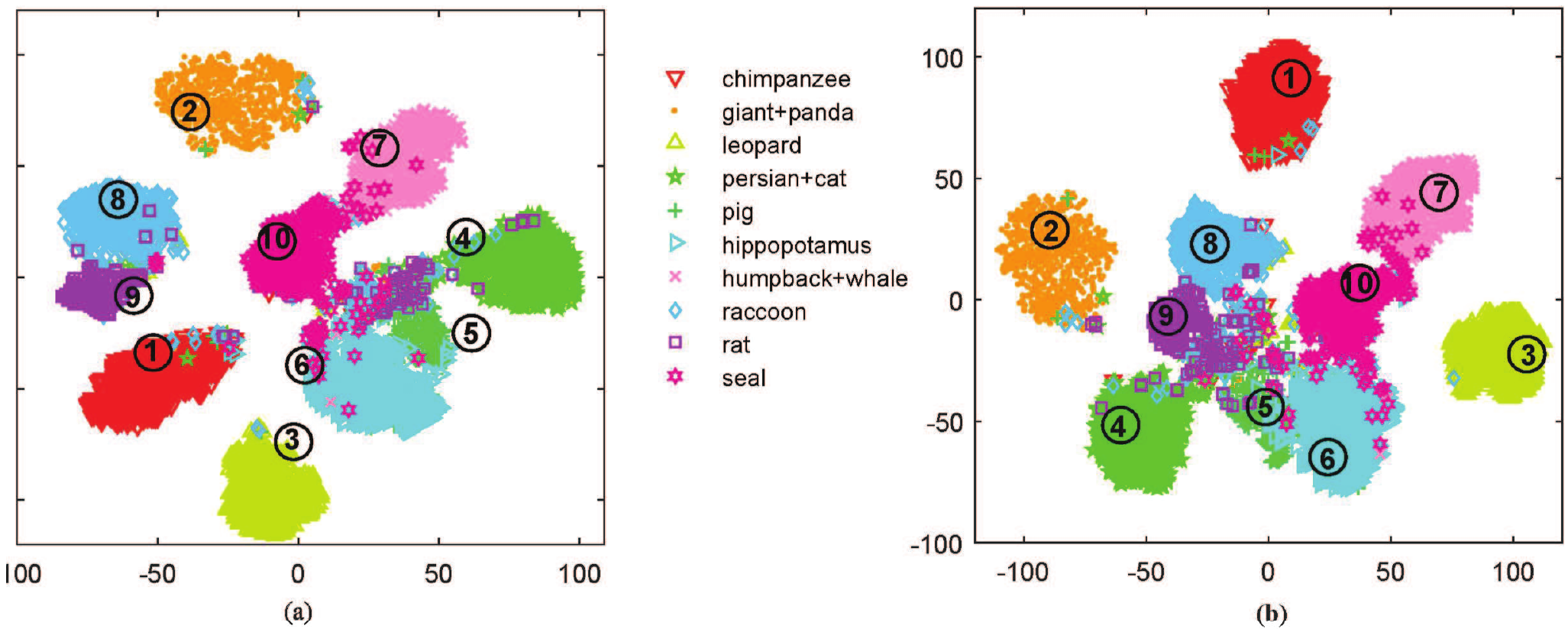}
\end{center}
   \caption{t-SNE visualization of unseen instances from AwA dataset. The black circles indicate the reconstructed visual prototypes with the corresponding class semantic descriptors. (a) denotes the visualization without normalization constraint while (b) denotes the visualization with normalization constraint. Best view in color.  }
\label{fig:long}
\label{fig:onecol}
\end{figure}

\begin{figure}[t]
\begin{center}
   \includegraphics[width=0.98\linewidth]{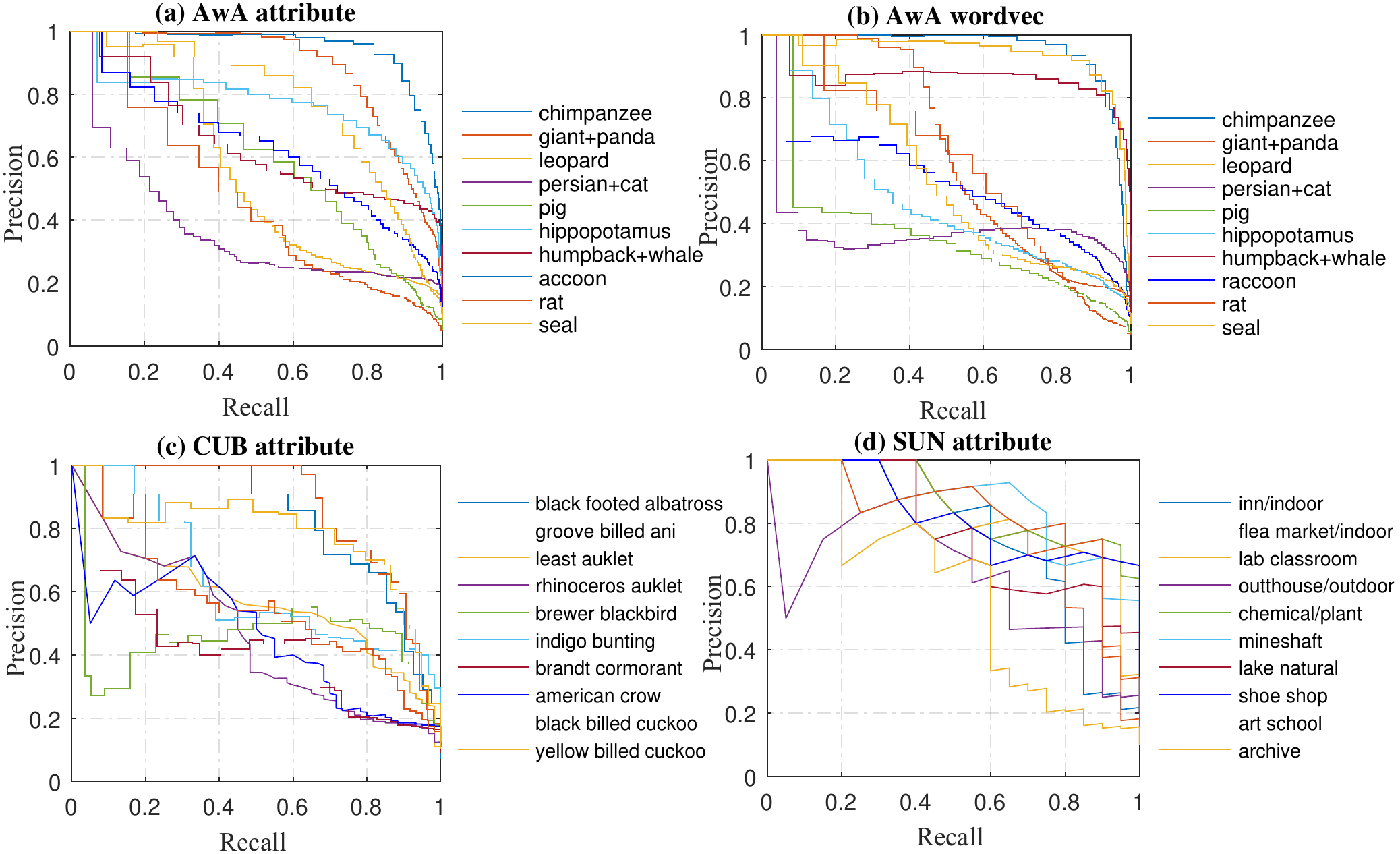}
\end{center}
   \caption{Precision-Recall curves for unseen classes on three datasets. For AwA, we plot the curves with attribute (a) and word vectors (b), respectively. For CUB dataset, we show the first 10 classes from 50 unseen classes. Best viewed in color.}
\label{fig:long}
\label{fig:onecol}
\end{figure}

\begin{table}
\begin{center}
\caption{\upshape{Zero-shot retrieval mAP (\%) comparison on three benchmark datasets. The results of the selective comparative methods are cited from the original papers.}}
\begin{tabular}{|l|c|c|c|c|}
\hline
Method & AwA &CUB & SUN & Ave. \\
\hline\hline
SSE \cite{Zhang15iccv} & 46.25 & 4.69 & 58.94 & 36.62\\
JLSE \cite{Zhang16cvpr} & 67.66 & 29.15 & 80.01 & 58.94\\
MLZSC \cite{Bucher16eccv} & 68.1 & 25.33 & 52.68 & 48.69\\
\hline
SSL-ZSL & \textbf{73.62} & \textbf{44.94} & \textbf{86.15} & \textbf{68.23} \\
\hline
\end{tabular}
\end{center}
\end{table}

\section{Conclusion}
We have proposed an end-to-end approach for zero-shot learning in which the semantic descriptors are embedded into the softmax layer in a multi-class classification framework. To narrow the structural differences between different modalities, an $L_2$ normalization constraint is introduced to imposed on the differences of visual features and the visual prototypes reconstructed with class semantic descriptors. We have shown experimentally that our method outperforms the state-of-the-arts methods both for zero-shot classification and zero-shot retrieval on AwA, CUB and SUN datasets, respectively.



%


\ifCLASSOPTIONcaptionsoff
  \newpage
\fi

\end{document}